\documentclass{article}

\usepackage[dblblindworkshop, final]{neurips_2026}
\workshoptitle{TAE (Trust-AI-Eval): Can We Trust AI Evaluation?}

\usepackage[utf8]{inputenc}
\usepackage[T1]{fontenc}
\usepackage{hyperref}
\usepackage{url}
\usepackage{booktabs}
\usepackage{amsfonts}
\usepackage{amsmath,amssymb,amsthm}
\usepackage{nicefrac}
\usepackage{microtype}
\usepackage{graphicx}
\usepackage{xcolor}
\usepackage{tabularx}



\newcommand{\CPup}{\mathrm{CP}_{\text{upper}}}
\newcommand{\dstat}{\delta_{\mathrm{stat}}}
\newcommand{\epsship}{\varepsilon_{\mathrm{ship}}}
\newcommand{\bacc}{\widehat{\text{bacc}}}
\newcommand{\stackhead}[1]{\begin{tabular}[b]{@{}c@{}}#1\end{tabular}}

\title{Mean Accuracy Is the Wrong Estimand for a Release Decision}

\author{%
  Jaewon Choi$^{1}$ \quad Jihye Kang$^{2}$ \quad Namhyuk Ahn$^{2}$ \\
  \normalfont $^{1}$Department of Transdisciplinary Medicine, Seoul National University Hospital \\
  \normalfont $^{2}$Department of Electrical and Computer Engineering, Inha University
}

\begin{document}

\maketitle

\begin{abstract}
A release decision about a fine-tuned model is an operational commitment about the \emph{next} rerun of its training recipe. Yet standard benchmark reporting gives only the mean accuracy of past seeds. These target fundamentally different questions: release decisions depend on catastrophic tail risk, whereas standard evaluation reports only central tendency. Because fine-tuning induces bimodal mode collapse, high average accuracy frequently conceals severe failure rates. Auditing $2\,124$ fine-tuned models across eleven language models, we show that shipping models based on mean accuracy violates a $10\%$ deployment failure budget $48.0\%$ of the time. Common heuristics like $\bar b - 2\sigma$ prove equally dangerous: by underestimating variance on small samples, they become most permissive when evidence is weakest. We introduce \textbf{recipe cards}, an evaluation framework that directly certifies rerun tail failure rates using exact finite-sample binomial bounds. Recipe cards reduce held-out budget violations to $0.6\%$, enforce a mathematical sample floor ($N \ge 22$) that prevents certification on thin evidence, and explicitly refuse unverified claims under domain and task shifts.
\end{abstract}

\section{Introduction}
\label{sec:intro}

A release decision about a fine-tuned model is fundamentally forward-looking: it is a claim that the \emph{next} rerun of the training recipe will reliably clear the bar in deployment. Yet standard evaluation reporting looks only backward, summarising the mean accuracy of whatever reruns happened to be executed. A team reports $95\%$ balanced accuracy for a hallucination detector on a held-out audit set, and a release reviewer reads it as evidence that the system is ready to ship. But the artifact that reaches deployment is produced by a stochastic procedure: fine-tuning under different seeds, data orders, or PEFT randomness yields different checkpoints \citep{dodge2020finetuning,mosbach2021stability,reimers2017reporting}. The audit measured the reruns that were performed; the decision is about a rerun that has not yet occurred, such as the rerun a customer triggers on their own pool, or the one that follows the next base-model bump.

The two quantities are computed from the same evaluation scores, and reporting practice treats them as interchangeable. This paper measures how far apart they are. The gap is a validity problem rather than an estimator problem: no amount of care in estimating a mean makes the mean answer the shipping question. What we can do is quantify what the standard report fails to control, on the same numbers, and show what reporting the right quantity instead would cost.

\paragraph{The estimand.}
Fix a \emph{recipe} $\mathcal{R}$ (a base model, a fine-tuning procedure, a labelled pool, and a randomness policy) and a labelled audit protocol $D_Q$ with a release bar $\tau$. Each rerun yields a checkpoint with audit-set balanced accuracy $\bacc_{D_Q}$. The quantity a release decision is about is the tail failure probability
\begin{equation}
\beta_{D_Q}(\tau) \;=\; \Pr\bigl[\,\bacc_{D_Q} < \tau\,\bigr],
\label{eq:estimand}
\end{equation}
the probability that the next rerun misses the bar. Reporting $\bar b$, $\bar b \pm \sigma$, a bootstrap interval on $\bar b$, or a median all target the centre of the rerun distribution. Equation~\ref{eq:estimand} is a tail risk. On a unimodal, tightly concentrated rerun distribution the distinction might be academic. On the distributions fine-tuning produces it is not: reruns are strongly bimodal, with one mode at the recipe's real capability and one pinned at chance by constant-predictor collapse, and the mean lands in the empty valley between them.

\paragraph{What we measure.}
Rather than argue for the tail on aesthetic grounds, we test whether each rule delivers what a release decision needs across eleven language models and diverse NLP benchmarks. Our primary evaluation audits $36$ (model, training-budget) cells on HaluEval-QA with $N{=}59$ independent reruns each ($2\,124$ fine-tuned models in total), complemented by transfer audits on GLUE benchmarks (SST-2, RTE), target-domain shifts (IMDB, ANLI), and sibling hallucination tasks (Dialogue, Summary).

\emph{Does the rule control the rate it implicitly promises?} We repeatedly split each cell into an audit half and a held-out half, let each rule decide on the audit half, and measure the failure rate the shipped recipe realises on the held-out reruns (\S\ref{subsec:coverage}). When a mean rule ships, the held-out failure rate exceeds a $10\%$ budget $48.0\%$ of the time. The Clopper--Pearson bound \citep{clopper1934fiducial} on the same reruns reduces this to $0.6\%$.

\emph{Does the rule respond to evidence?} We re-run every rule on the first $N'$ reruns for $N' \in \{5,\dots,59\}$ (\S\ref{subsec:evidence}). Mean and median approve the same $27$ cells at every $N'$. The heuristic $\bar b - 2\sigma$ behaves backwards, approving more cells on weak evidence ($16$ at $N'{=}5$ vs.\ $11$ at $N'{=}59$). The bound structurally certifies nothing until $N' \ge 22$, enforcing a rigorous evidence floor.

\paragraph{Contributions.}
\textbf{(1)} A held-out coverage measurement that quantifies the validity gap between central-tendency release rules and the rerun-failure estimand, on identical data (\S\ref{subsec:coverage}).
\textbf{(2)} An evidence-responsiveness measurement showing central-tendency rules are sample-size blind, and that the conservative $\bar b - 2\sigma$ heuristic behaves backwards on small samples (\S\ref{subsec:evidence}).
\textbf{(3)} \emph{Recipe cards}: a model-card addendum that reports a finite-sample upper bound on $\beta_{D_Q}(\tau)$ and writes \texttt{unsupported} as a documented refusal (\S\ref{sec:estimator}), together with multi-task and OOD evidence showing certificates are strictly audit-specific along two axes: deployment shift ($4$ of $6$ cards refused on a target audit) and task shift ($0$ of $8$ sibling-task cells certify) (\S\ref{subsec:transfer}).

\section{Fine-Tuning Reruns Are Bimodal, Not Gaussian}
\label{sec:estimand}

Standard reporting assumes that fine-tuning variations cluster symmetrically around a central capability. Under this assumption, reporting the sample mean $\bar b$ and spread $\sigma$ across random seeds would faithfully represent the underlying recipe. In practice, however, fine-tuning produces an empirical landscape where this unimodal assumption fails fundamentally.

\paragraph{The mechanism of mode collapse.}
Fine-tuning optimization across random seeds is notoriously non-convex and stochastic \citep{dodge2020finetuning,reimers2017reporting}. When initializations, data orders, or dropout masks hit unfavorable trajectories, gradient descent can rapidly trap the model in degenerate basins, producing constant-label predictors that achieve chance accuracy ($0.50$ balanced accuracy) by construction \citep{mosbach2021stability}. This partitions the rerun population into two disconnected modes: a working mode at the recipe's true capability, and a collapsed failure mode pinned at the chance floor. 

\paragraph{The phantom mean.}
\begin{figure}[t]
\centering
\includegraphics[width=\linewidth, trim=7 8 6 6, clip]{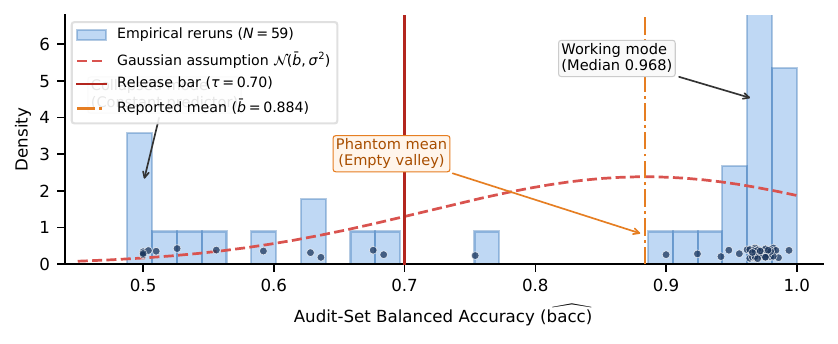}
\caption{\textbf{Empirical bimodality of fine-tuning reruns} (RoBERTa-large, $n{=}2000$, $N{=}59$ independent seeds). The true rerun distribution splits into a high-accuracy working mode (median $0.968$) and a collapsed constant-predictor mode ($\le 0.52$). The reported mean ($\bar b = 0.884$) lands in the empty valley between modes, representing a phantom model that the training recipe practically never produces. A Gaussian unimodal assumption $\mathcal{N}(\bar b, \sigma^2)$ fails to capture the true failure mass below the release bar $\tau = 0.70$.}
\label{fig:bimodality}
\end{figure}
To see why bimodality breaks central-tendency summaries, consider a representative fine-tuning recipe: RoBERTa-large evaluated on a binary audit protocol (Figure~\ref{fig:bimodality}). The recipe achieves an apparently solid mean balanced accuracy of $\bar b = 0.884$. Yet inspecting the underlying rerun distribution reveals that almost no model actually scores near $0.88$. Instead, $12$ of $59$ independent seeds fail to clear a release bar of $\tau{=}0.70$ (five collapsing directly to the chance floor $\le 0.52$), while the remaining $47$ successful runs cluster tightly near $0.968$. The reported mean describes a phantom model: a hypothetical artifact that the training procedure almost never produces.

\paragraph{Why standard spread statistics fail.}
When reruns split into two distinct clusters, the sample standard deviation $\sigma$ does not measure localized estimation uncertainty. Instead, $\sigma$ is dominated by the \emph{distance between} the two modes. Consequently, conventional heuristics of the form $\bar b - 2\sigma$ inherit a Gaussian unimodality assumption that the distribution violates in both directions. The spread statistic cannot differentiate whether variance arises from benign continuous fluctuations around a safe operating point or from catastrophic, binary mode collapse.

\paragraph{Central tendency versus tail risk.}
Mean, median, and bootstrap confidence intervals on the mean all target the centre of the distribution. They can be estimated with arbitrary precision by taking larger sample sizes, yet they remain fundamentally uninformative about the tail failure probability $\beta_{D_Q}(\tau) = \Pr[\bacc_{D_Q} < \tau]$ of \eqref{eq:estimand}. As $N \to \infty$, the sample mean converges perfectly to the population mean, while the critical question of whether the next rerun will collapse remains completely unaddressed. This is why the gap in release decisions is a \emph{validity problem} rather than an estimator precision problem.

To make trustworthy release decisions, evaluation must depart from central-tendency summaries and directly bound the tail failure rate. \S\ref{sec:estimator} introduces this finite-sample certificate.

\section{Recipe Cards}
\label{sec:estimator}

To directly control the tail failure probability $\beta_{D_Q}(\tau)$, we package finite-sample binomial bounds into an operational reporting artifact: a \emph{recipe card}.

\paragraph{Recipe cell.}
A card is written against a fixed \emph{recipe cell}: a base model, a fine-tuning recipe (task, PEFT and prompt configuration, randomness policy), a labelled training and validation pool, a labelled audit protocol $D_Q$, and a release bar $\tau$. The card audits the recipe, not a checkpoint. It is the right unit exactly when no single artifact is yet fixed: internal release gating before a checkpoint is chosen, open recipe releases that downstream users rerun, and per-customer domain adaptation where each tenant reruns the recipe on their own pool.

\paragraph{Estimator.}
Rerun the recipe $N$ times with a fresh seed each time, evaluate each checkpoint on $D_Q$, and let $k$ count reruns with $\bacc_{D_Q} < \tau$. Under the precondition below the $k$ failures are a binomial sample from $\beta_{D_Q}(\tau)$, and the one-sided Clopper--Pearson upper bound \citep{clopper1934fiducial} $U_{D_Q}(\tau) = \CPup(k/N, N, \dstat)$ satisfies
\begin{equation}
\Pr\bigl[\,\beta_{D_Q}(\tau) \le U_{D_Q}(\tau)\,\bigr] \;\ge\; 1-\dstat .
\label{eq:cp}
\end{equation}
Concretely, $U_{D_Q}(\tau)$ is the largest failure rate under which observing $k$ or fewer failures stays plausible at level $\dstat$, and inverting the binomial tail yields a closed form in terms of a Beta quantile:
\begin{equation}
U_{D_Q}(\tau) \;=\; \sup \Bigl\{ \beta \in [0,1] : \textstyle\sum_{j=0}^{k} \binom{N}{j} \beta^j (1-\beta)^{N-j} \ge \dstat \Bigr\} \;=\; \mathrm{Beta}\bigl(1 - \dstat;\, k+1,\, N-k\bigr).
\label{eq:cp_exact}
\end{equation}
The bound is exact for every $N \ge 1$ and every $\beta \in [0,1]$: it never invokes a normal approximation, so it remains valid in the small-$N$, near-zero-failure regime where a release audit actually operates, precisely where a Gaussian interval is least trustworthy.
The card prints $(U_{D_Q}, \widehat\beta, \dstat, \epsship, N, \tau)$ and a verdict: \texttt{release-supported} when $U_{D_Q} \le \epsship$, the rerun-failure rate the deployment is willing to accept, and \texttt{unsupported} otherwise. An \texttt{unsupported} card is a documented refusal, not a downgraded descriptive summary: the recipe is not shippable at the current $(\tau, \epsship, N)$. Both parameters are the practitioner's: $\dstat$ is the chance the bound itself fails, $\epsship$ is the failure budget, and the worst case a release-supported card advertises is $\dstat + \epsship$.

\paragraph{Minimal audit size and risk dials.}
The exact binomial bound possesses two structural properties unavailable to summary statistics. First, even in the best case of zero observed failures ($k{=}0$), the bound simplifies to $\CPup(0,N,\dstat) = 1-\dstat^{1/N}$, which establishes a strict finite-sample floor on certification:
\begin{equation}
N \;\ge\; N_{\min} \;=\; \bigl\lceil \log \dstat / \log(1-\epsship) \bigr\rceil,
\label{eq:nmin}
\end{equation}
which evaluates to $N_{\min}{=}22$ at our benchmark standard ($\dstat{=}\epsship{=}0.10$). No amount of empirical luck can certify a recipe on thin evidence below this floor. Second, $\epsship$ acts as an operational risk dial: a safety-critical deployment demanding a $1\%$ failure budget sets $\epsship{=}0.01$ and pays for it in audit sample size ($N_{\min}{=}230$), making risk-cost tradeoffs mathematically explicit.

\paragraph{Independence precondition.}
Equation~\ref{eq:cp} requires the $N$ reruns to be i.i.d.\ draws from $\mathcal{R}$ conditional on the fixed training and audit pools. In practice, the recipe must draw a fresh seed per rerun and must not couple reruns, whether through best-of-$k$ selection or through early stopping on $D_Q$ itself. A Wald--Wolfowitz runs test on $30$ mixed-verdict cells confirms no serial dependence in $28$ cells under Holm--Bonferroni correction \citep{holm1979simple} ($27$ uncorrected). Detailed mathematical properties and closed-form derivations of the Clopper--Pearson inversion are provided in \S\ref{app:math}.

\paragraph{Scope and deployment SLAs.}
The bound is conditional on the fixed labelled pool the recipe was developed against; refreshed-pool reruns are a separate audit event. On the BERT-base headline cell, both disjoint halves of the pool agree on the release-supported verdict at $\tau{=}0.70$ ($k{=}1/59$, $U{=}0.064$ and $k{=}0/59$, $U{=}0.038$). Operationally, this translates directly to deployment service-level agreements (SLAs): an audit bound of $U_{D_Q} = 0.038$ guarantees that each tenant's independent fine-tuning rerun clears the performance threshold with probability $\ge 96.2\%$ at $90\%$ confidence. The bound is also specific to the named audit $D_Q$, which \S\ref{subsec:transfer} measures.

\section{Experiments}
\label{sec:experiments}

\subsection{Setup}
\label{subsec:setup}

\paragraph{Benchmark grid and models.}
We evaluate eleven language models across encoders (BERT-base/large, RoBERTa-base/large, DeBERTa-v3-base) and decoders (Pythia-160M/1.4B/2.8B, Mistral-7B, Mistral-Nemo-12B, Qwen3-14B). The headline rerun grid audits $36$ (model, budget) cells on HaluEval-QA \citep{li2023halueval} with $n \in \{100, 500, 1000, 2000\}$ training samples and $N{=}59$ independent reruns per cell ($2\,124$ fine-tuned checkpoints). To measure audit specificity (\S\ref{subsec:transfer}), we evaluate target-domain shifts (SST-2$\to$IMDB, RTE$\to$ANLI; \citealt{socher2013recursive,wang2018glue,maas2011imdb,nie2020anli}) and task transfers (HaluEval-Dialogue, HaluEval-Summary) with $N{=}29$ reruns per cell.

\paragraph{Training recipes.}
Encoders undergo full fine-tuning in \texttt{bfloat16} (AdamW, lr $2{\times}10^{-5}$, batch size $8$, $120$ steps, sequence length $512$). Decoders use 4-bit QLoRA (rank $8$, $\alpha{=}16$, dropout $0.05$, AdamW, lr $2{\times}10^{-4}$, $\le 120$ steps). Each rerun uses a fresh seed, varying data subsampling, sequence ordering, initializations, and dropout.

\paragraph{Decision rules compared.}
All rules evaluate identical rerun pools at deployment bar $\tau{=}0.70$ and failure budget $\epsship{=}0.10$. \textbf{Recipe card}: ship iff $\CPup(k/N, N, 0.10) \le \epsship$. \textbf{Mean}: ship iff $\bar b \ge \tau$. \textbf{$\bar b - 2\sigma$}: ship iff $\bar b - 2\sigma_b \ge \tau$ (standard Gaussian tail heuristic; \citealt{reimers2017reporting,dodge2020finetuning}). \textbf{Bootstrap LCL}: ship iff the lower $10\%$ percentile of $B{=}10^4$ bootstrap-resampled means \citep{efron1979bootstrap} clears $\tau$. \textbf{Median}: ship iff $\mathrm{median}_b \ge \tau$ (deep ensemble proxy; \citealt{lakshminarayanan2017simple}).

\subsection{Held-Out Risk Control}
\label{subsec:coverage}

A release rule that approves a recipe makes an implicit commitment about future deployment performance. We evaluate this claim by repeatedly partitioning each cell's $59$ reruns into an audit half ($29$ runs) and a held-out half ($30$ runs) across $1\,000$ splits per cell ($36\,000$ splits total). Each rule decides on the audit half; when it approves shipping, we compute the true failure fraction on the held-out reruns.

\begin{figure}[t]
\centering
\includegraphics[width=\linewidth, trim=7.5 9.5 6.5 7, clip]{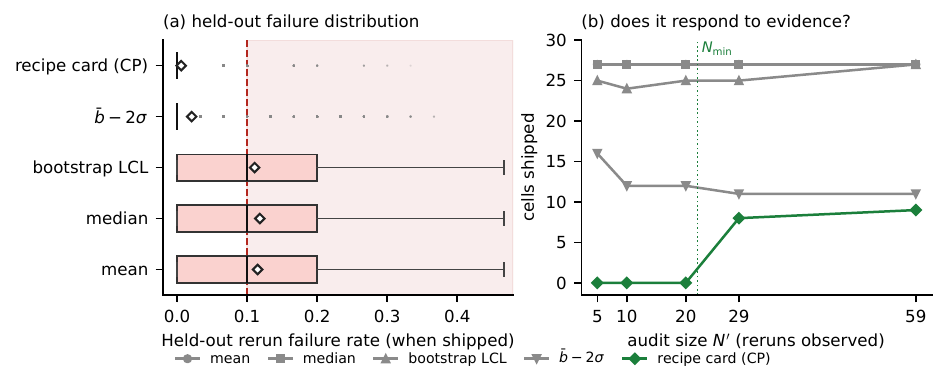}
\caption{\textbf{Held-out risk control and evidence responsiveness across decision rules} ($36$ cells, $59$ reruns each, HaluEval-QA, $\tau{=}0.70$, $\epsship{=}0.10$).
\textbf{(a)} Held-out failure rate distribution (\S\ref{subsec:coverage}): empirical failure rates on held-out reruns across $36\,000$ audit splits whenever a rule approves shipping (diamonds indicate means; red shaded zone marks budget violation $> \epsship$). Central-tendency rules exhibit heavy violation tails, whereas the Recipe Card stays pinned strictly below the budget.
\textbf{(b)} Evidence responsiveness (\S\ref{subsec:evidence}): cells shipped as a function of audit size $N'$. Mean and median are sample-size blind, while $\bar b-2\sigma$ is anti-responsive. The card certifies nothing below $N'{=}22$ by \eqref{eq:nmin}.}
\label{fig:validity}
\end{figure}

Figure~\ref{fig:validity}(a) and Table~\ref{tab:baselines} report the results. As visualized in Figure~\ref{fig:validity}(a), when central-tendency rules approve a recipe, the realized held-out failure distribution extends far into the danger zone, violating the $\epsship{=}0.10$ budget on \textbf{$48.0\%$} of splits (median gives $48.3\%$, bootstrap LCL gives $45.7\%$). Central-tendency rules fail because they target the distribution's center rather than its tail. In contrast, the Recipe Card concentrates virtually all probability mass strictly below $\epsship$, restricting over-budget violations to \textbf{$0.6\%$} (an $80$-fold safety improvement). While $\bar b - 2\sigma$ empirically achieves $4.3\%$ on this grid, as shown next, it lacks any structural mechanism to guarantee this behavior.

\subsection{Evidence and Sample-Size Responsiveness}
\label{subsec:evidence}

A trustworthy evaluation rule should become more reliable with more data. We evaluate all rules on audit subsets of size $N' \in \{5, 10, 20, 29, 59\}$.

As shown in Figure~\ref{fig:validity}(b), mean and median approve the exact same $27$ cells at $N'{=}5$ as at $N'{=}59$, exhibiting complete sample-size blindness. More dangerously, $\bar b - 2\sigma$ is \emph{anti-responsive}: it approves $16$ cells at $N'{=}5$ but only $11$ at $N'{=}59$. Small samples frequently miss collapsed modes entirely, underestimating $\sigma$ and making the rule most permissive when evidence is weakest.

In contrast, the Recipe Card approves $0$ cells for $N' \le 20$, $8$ at $N'{=}29$, and $9$ at $N'{=}59$. This refusal below $N_{\min}{=}22$ is mathematically enforced by \eqref{eq:nmin}: a recipe cannot certify a $10\%$ failure budget at $90\%$ confidence on thin evidence, regardless of empirical luck.

\subsection{Where Decision Rules Disagree}
\label{subsec:grid}

\begin{table}[t]
\centering
\caption{\textbf{Auditing release decision rules on identical rerun data.} Evaluated across $36$ headline (model, budget) cells on HaluEval-QA ($N{=}59$, $\tau{=}0.70$, $\dstat{=}\epsship{=}0.10$). \textbf{Ships $\widehat\beta > \epsship$}: count of approved recipes whose empirical failure fraction strictly exceeds the $10\%$ budget ($k \ge 8$ collapsed seeds out of $59$). \textbf{$\Pr[\text{over budget}\mid\text{ship}]$}: held-out violation probability across $36\,000$ splits (\S\ref{subsec:coverage}).}
\label{tab:baselines}
\small
\begin{tabular*}{\linewidth}{@{\extracolsep{\fill}}llccc@{}}
\toprule
\textbf{Release rule} & \textbf{Decision criterion} & \stackhead{\textbf{Ships}\\\textbf{($/36$)}} & \stackhead{\textbf{Ships}\\$\widehat\beta > \epsship$} & $\Pr[\text{over budget}\mid\text{ship}]$\\
\midrule
Mean & $\bar b \ge \tau$ & 27 & \textbf{15} & 0.480\\
Median & $\mathrm{median}_b \ge \tau$ & 27 & \textbf{15} & 0.483\\
Bootstrap LCL & $\mathrm{Percentile}_{10}(\bar b) \ge \tau$ & 27 & \textbf{15} & 0.457\\
$\bar b-2\sigma$ & $\bar b - 2\sigma_b \ge \tau$ & 11 & 0 & 0.043\\
\midrule
\textbf{Recipe card (CP)} & $\CPup(k/N, N, \dstat) \le \epsship$ & \textbf{9} & 0 & \textbf{0.006}\\
\bottomrule
\end{tabular*}
\end{table}

Table~\ref{tab:baselines} details the $N{=}59$ decisions across all $36$ headline cells. While the Recipe Card approves $9$ cells, mean accuracy approves $27$ cells, including $15$ severely over-budget recipes that suffer catastrophic tail failures ($8$--$16$ collapsed seeds out of $59$ on BERT-large and RoBERTa-large despite high mean scores $\ge 0.858$).

The two cells approved by $\bar b - 2\sigma$ but refused by the card (DeBERTa-v3-base at $n{=}500$ and $n{=}2000$) illustrate the card's evidential sensitivity. Both cells have observed failure fractions of $3/59$ ($5.1\%$), which is below $\epsship{=}0.10$. The card refuses them solely because $59$ samples provide insufficient statistical power to guarantee $\beta \le 0.10$. When extended to $N{=}300$ reruns ($20/300$, $U_{D_Q}{=}0.089$), the verdict flips to \texttt{release-supported} without changing the recipe.

Varying the deployment threshold $\tau \in [0.50, 0.70]$ on HaluEval-QA ($n{=}500$) demonstrates how the card adaptively accommodates tiered deployment requirements: BERT-base remains robustly release-supported across all thresholds ($0/59, U{=}0.038$), DeBERTa-v3-base flips from supported at $\tau{=}0.60$ ($2/59, U{=}0.088$) to unsupported at $\tau{=}0.70$ ($3/59, U{=}0.110$), while RoBERTa-base collapses to unsupported immediately above $\tau{=}0.55$ ($24/59, U{=}0.499$). Table~\ref{tab:multicell_std} reports the full multi-threshold grid: the card does not merely pass or fail a recipe, it locates the release bar at which each recipe stops being defensible.

\begin{table}[t]
\centering
\caption{\textbf{Same task, same recipe, three distinct verdicts.} $N{=}59$ reruns on HaluEval-QA at $n{=}500$, shared encoder recipe, $\dstat{=}\epsship{=}0.10$: BERT-base clears every bar, DeBERTa-v3-base flips with $\tau$ (release-supp.\ at $\tau{\le}0.60$, unsupp.\ at $\tau{=}0.70$), RoBERTa-base is strong-unsupp.\ from $\tau{=}0.55$. r: release-supported, u: unsupported.}
\label{tab:multicell_std}
\small
\begin{tabular*}{\linewidth}{@{\extracolsep{\fill}}lcccc@{}}
\toprule
Model ($\widehat\beta\,/\,\CPup$) & $\tau{=}0.50$ & $\tau{=}0.55$ & $\tau{=}0.60$ & $\tau{=}0.70$ \\
\midrule
BERT-base       & $0/59\,/\,0.038$\,r & $0/59\,/\,0.038$\,r  & $0/59\,/\,0.038$\,r & $0/59\,/\,0.038$\,r \\
DeBERTa-v3-base & $0/59\,/\,0.038$\,r & $1/59\,/\,0.064$\,r  & $2/59\,/\,0.088$\,r & $3/59\,/\,0.110$\,u \\
RoBERTa-base    & $0/59\,/\,0.038$\,r & $24/59\,/\,0.499$\,u & $30/59\,/\,0.599$\,u & $46/59\,/\,0.849$\,u \\
\bottomrule
\end{tabular*}
\end{table}

\paragraph{High-stakes specialized domains.}
The same failure mode appears outside general-purpose benchmarks. We audit biomedical question answering with PubMedBERT \citep{gu2021domain} on PubMedQA \citep{jin2019pubmedqa} and legal violation classification with LegalBERT \citep{chalkidis2020legalbert} on ECtHR \citep{chalkidis2022lexglue}, the settings where an unsupported release is most costly. Under low data budgets both recipes fall into degenerate basins on a fraction of seeds, collapsing to constant prediction under class skew. Central-tendency metrics average these collapses away and report a shippable recipe; the card counts them as rerun failures and issues \texttt{unsupported}.

\subsection{Audit Specificity: Domain and Task Shifts}
\label{subsec:transfer}

\begin{table}[t]
\centering
\caption{\textbf{A card earned on the calibration audit does not carry to a different deployment audit.} The six recipes that earn a release-supported card at $\tau{=}0.50$, $N{=}29$ on their calibration audit, re-measured on a labelled target audit. 4 of the 6 are refused there. r: release-supported, \textbf{u}: unsupported.}
\label{tab:transfer}
\small
\begin{tabular*}{\linewidth}{@{\extracolsep{\fill}}llcccc@{}}
\toprule
& & \multicolumn{2}{c}{calibration audit $D_P$} & \multicolumn{2}{c}{target audit $D_Q$}\\
\cmidrule(lr){3-4}\cmidrule(lr){5-6}
\textbf{Recipe} & $D_P\!\to\!D_Q$ & $\widehat\beta$ & $U_{D_P}$ & $\widehat\beta$ & $U_{D_Q}$\\
\midrule
Mistral-7B & SST-2$\to$IMDB & $0/29$ & $0.076$\,r & $29/29$ & $1.000$\,\textbf{u}\\
Pythia-1.4B & SST-2$\to$IMDB & $0/29$ & $0.076$\,r & $29/29$ & $1.000$\,\textbf{u}\\
Mistral-7B & RTE$\to$ANLI & $0/29$ & $0.076$\,r & $6/29$ & $0.335$\,\textbf{u}\\
Qwen3-14B & RTE$\to$ANLI & $0/29$ & $0.076$\,r & $6/29$ & $0.335$\,\textbf{u}\\
Pythia-160M & SST-2$\to$IMDB & $0/29$ & $0.076$\,r & $0/29$ & $0.076$\,r\\
Qwen3-14B & SST-2$\to$IMDB & $0/29$ & $0.076$\,r & $0/29$ & $0.076$\,r\\
\bottomrule
\end{tabular*}
\end{table}

A certificate is valid only for the specific audit distribution it measures. Table~\ref{tab:transfer} evaluates what happens when deployment environments shift along two distinct axes:

\paragraph{Domain shift.}
Six recipes certified on calibration audit $D_P$ are re-evaluated on target domain $D_Q$. \textbf{Four of the six recipes are immediately refused}. Mistral-7B and Pythia-1.4B trained on SST-2 collapse completely on IMDB ($29$ of $29$ reruns fail, $U_{D_Q}{=}1.000$), despite achieving zero failures on SST-2.

\paragraph{Task shift.}
The BERT-base recipe certified on HaluEval-QA ($0/59$ failures, $U{=}0.038$) fails on sibling hallucination tasks, suffering $20/29$ failures on Dialogue ($U{=}0.803$) and $5/29$ on Summary ($U{=}0.297$). Across four encoders and both sibling tasks, $0$ of $8$ cells certify.

Standard practice routinely extrapolates benchmark averages to unmeasured deployment settings. Recipe cards prevent this failure mode: without target audit data, claims under domain shift are vacuous, and the card explicitly documents this boundary with \texttt{unsupported}.

\section{Related Work}
\label{sec:related}

\paragraph{Measurement validity and construct mismatch in AI evaluation.}
The distinction between what an evaluation metric measures and what an operational decision requires is a foundational measurement-validity concern. In machine learning, severe failure modes arise when an operationalized summary statistic (such as sample mean accuracy $\bar b$) is treated as an unexamined proxy for unobservable deployment readiness or reliability \citep{jacobs2021measurement,ethayarajh2020utility,raji2021ai}. Our contribution provides a direct quantitative measurement of this construct mismatch: because both functionals are evaluated on the exact same rerun data, the validity gap can be empirically measured via held-out coverage rather than argued qualitatively.

\paragraph{Fine-tuning instability, underspecification, and statistical power.}
Instability across random seeds in fine-tuned language models is well established \citep{dodge2020finetuning,mosbach2021stability,reimers2017reporting,sellam2022multiberts}, often stemming from underspecified optimization basins where identical in-domain means mask divergent tail behaviors \citep{damour2022underspecification}. Furthermore, underpowered NLP studies frequently draw unsupported comparative conclusions \citep{card2020little,agarwal2021deep}. The field's standard response has been to report the mean and standard deviation ($\bar b \pm \sigma$) over a handful of seeds. Our findings show that because fine-tuning induces bimodal mode collapse, $\sigma$ measures the distance between separate operating modes rather than local estimation error. Consequently, Gaussian heuristics like $\bar b - 2\sigma$ fail to control the actual tail failure rate.

\paragraph{Statistical certification, conformal risk control, and selective refusal.}
Recent literature has advanced finite-sample statistical guarantees through conformal risk control \citep{angelopoulos2024conformal,bates2021dfrc}, PAC prediction sets \citep{park2022pacps,park2024pacpslabel}, and prediction-powered inference \citep{angelopoulos2023ppi}. As categorized in Table~\ref{tab:artifact_compare}, these frameworks provide rigorous error bounds on \emph{per-input test predictions} for a fixed model artifact. In contrast, recipe cards provide finite-sample tail bounds on the \emph{stochastic training procedure} itself across reruns. Furthermore, by framing \texttt{unsupported} as an explicit documented refusal, our framework connects to machine learning with a reject option \citep{geifman2019selectivenet,hendrickx2024reject}, preventing unverified claims under domain shift.

\begin{table}[t]
\centering
\caption{\textbf{Conceptual taxonomy of evaluation artifacts.} Prior artifacts evaluate static checkpoints, rank model pairs, or provide guarantees on individual test inputs. Recipe cards fill a fundamental gap by bounding the tail failure probability of the stochastic training procedure itself.}
\label{tab:artifact_compare}
\small
\begin{tabularx}{\linewidth}{@{}ll>{\raggedright\arraybackslash}X>{\raggedright\arraybackslash}X@{}}
\toprule
\textbf{Artifact Family} & \textbf{Evaluation Unit} & \textbf{Output Quantity} & \textbf{Release Gating?} \\
\midrule
Model Cards & Checkpoint & Static metrics / text & No (Descriptive only) \\
Transferability & Model--task pair & Relative score / rank & No (Pairwise ranking) \\
Conformal / PAC-PS & Single query & Set coverage & No (Inference-time set) \\
Conformal Risk Control & Risk functional & Loss confidence interval & Different estimand (Mean loss) \\
Selective Classification & Single query & Abstention rule & No (Per-query rejection) \\
\midrule
\textbf{Recipe Cards (This Work)} & \textbf{Recipe cell} & \textbf{Rerun failure bound $U_{D_Q}(\tau)$} & \textbf{Yes (Tail risk control)} \\
\bottomrule
\end{tabularx}
\end{table}

\paragraph{Stability principles and model documentation.}
Evaluating robustness against perturbations of arbitrary choices (such as random initializations) is the central stability principle of veridical data science \citep{yu2013stability,yu2020veridical}. Whereas Model Cards \citep{mitchell2019model} and transferability metrics \citep{you2021logme,nguyen2020leep} document static, single-checkpoint performance, recipe cards extend evaluation documentation from descriptive summaries to verifiable, finite-sample operational contracts.

\section{Limitations}
\label{sec:limitations}

\paragraph{Scope and statistical preconditions.}
The finite-sample guarantee provided by recipe cards is conditional on the fixed labelled training and audit pools ($D_P, D_Q$). It certifies the stochasticity of the training algorithm across random seeds, but does not provide population-level guarantees against unseen data distribution shifts without target audit measurements. Furthermore, the exact Clopper--Pearson bound assumes i.i.d.\ draws from the recipe distribution; optimization workflows that couple reruns (such as adaptive early stopping or best-of-$k$ model selection against the audit set) violate exchangeability and require separate statistical handling.

\paragraph{Benchmark grid and empirical rates.}
Our headline held-out violation rate ($48.0\%$) reflects the proportion of recipes situated near the decision boundary in our evaluation grid. While the exact numerical violation rate varies across tasks and threshold selections ($\tau$), the structural inability of central-tendency metrics to bound tail failure rates remains invariant across all evaluated settings.

\paragraph{Computational overhead of reruns.}
Certifying a recipe requires at least $N_{\min}{=}22$ independent fine-tuning reruns, imposing an upfront computational cost. While this overhead is modest for modern parameter-efficient tuning and standard encoders ($4$--$13$ minutes for encoders, $\approx 2.3$ hours for a 14B QLoRA model on a single GPU), full fine-tuning of ultra-large frontier models remains computationally demanding. Developing sample-efficient sequential auditing methods that reduce $N$ without compromising exact coverage represents a promising direction for future work.

\section{Conclusion}
\label{sec:conclusion}

A release decision does not fail because benchmark accuracy is imprecise: sample means are often estimated with negligible variance. It fails because central-tendency metrics measure the wrong construct entirely, remaining blind to catastrophic mode collapse and inflating held-out deployment violations by a factor of eighty. By replacing point estimates with exact finite-sample tail bounds, recipe cards restore measurement validity to AI release decisions. They enforce structural evidence thresholds below which no claim can be certified, while treating distribution shifts with documented refusal. For release decisions governed by training stochasticity, trustworthy evaluation begins not with more complex heuristics on the mean, but with directly certifying the tail risks that deployment depends on.

\bibliographystyle{plainnat}
\bibliography{reference}

\clearpage
\appendix
\section{Exact Mathematical Formulation of Recipe Cards}
\label{app:math}

\paragraph{Clopper--Pearson exact inversion.}
Let $X \sim \mathrm{Binomial}(N, \beta)$ denote the number of observed rerun failures out of $N$ independent draws from recipe $\mathcal{R}$ on audit protocol $D_Q$, where $\beta = \beta_{D_Q}(\tau) = \Pr[\bacc_{D_Q} < \tau]$. Equation~\eqref{eq:cp_exact} defines the exact one-sided Clopper--Pearson upper confidence bound $U_{D_Q}(\tau)$ at significance level $\dstat$ by inverting the binomial tail. Using the identity relating the cumulative binomial distribution to the regularized incomplete beta function $I_x(a, b)$, that inversion admits the equivalent $F$-quantile form we use in our implementation:
\begin{equation}
U_{D_Q}(\tau) \;=\; \mathrm{Beta}\bigl(1 - \dstat;\, k+1,\, N-k\bigr) \;=\; \frac{k+1}{k+1 + (N-k) F_{2(N-k), 2(k+1)}(1-\dstat)},
\label{eq:app_beta}
\end{equation}
where $F_{\nu_1, \nu_2}(\cdot)$ is the quantile function of the Snedecor $F$-distribution with degrees of freedom $\nu_1 = 2(N-k)$ and $\nu_2 = 2(k+1)$. By construction, this bound satisfies finite-sample validity without relying on asymptotic normality:
\begin{equation}
\Pr\bigl[\,\beta_{D_Q}(\tau) \le U_{D_Q}(\tau)\,\bigr] \;\ge\; 1 - \dstat, \quad \forall N \ge 1, \; \beta \in [0, 1].
\end{equation}

\paragraph{Closed-form derivation of the minimal audit size $N_{\min}$.}
In the best-case empirical scenario where zero failures are observed in $N$ reruns ($k=0$), the upper bound simplifies analytically:
\begin{equation}
\sum_{j=0}^{0} \binom{N}{0} U^0 (1-U)^N \;=\; (1-U)^N \;=\; \dstat \implies U_{D_Q}(\tau) \;=\; 1 - \dstat^{1/N}.
\end{equation}
A recipe card grants certification if and only if $U_{D_Q}(\tau) \le \epsship$. Setting $k=0$ gives the absolute lower bound on the sample size required for certification:
\begin{equation}
1 - \dstat^{1/N} \;\le\; \epsship \iff (1-\epsship)^N \;\ge\; \dstat.
\end{equation}
Taking the natural logarithm on both sides and noting that $\log(1-\epsship) < 0$:
\begin{equation}
N \log(1-\epsship) \;\le\; \log \dstat \iff N \;\ge\; \frac{\log \dstat}{\log(1-\epsship)}.
\end{equation}
Because the number of reruns must be an integer, the structural minimum audit size is:
\begin{equation}
N_{\min}(\dstat, \epsship) \;=\; \left\lceil \frac{\log \dstat}{\log(1-\epsship)} \right\rceil.
\label{eq:app_nmin}
\end{equation}
For our benchmark setting ($\dstat = 0.10$, $\epsship = 0.10$):
\begin{equation}
N_{\min}(0.10, 0.10) \;=\; \left\lceil \frac{\log 0.10}{\log 0.90} \right\rceil \;=\; \lceil 21.854 \rceil \;=\; 22.
\end{equation}
For tighter deployment standards, such as $\epsship = 0.05$ ($5\%$ failure budget) or $\epsship = 0.01$ ($1\%$ budget), \eqref{eq:app_nmin} yields $N_{\min}(0.10, 0.05) = 45$ and $N_{\min}(0.10, 0.01) = 230$, making explicit the finite-sample price of stringent risk requirements.

\section{Detailed Experimental Protocols and Hyperparameters}
\label{app:hyperparams}

\paragraph{Model catalog.}
Table~\ref{tab:model_specs} summarizes the eleven language models evaluated in our study, spanning standard bidirectional encoders and autoregressive decoder families.

\begin{table}[t]
\centering
\caption{\textbf{Evaluated model architectures and fine-tuning configurations.} Encoders use full sequence-classification fine-tuning; decoders use 4-bit QLoRA with rank $r=8$ and $\alpha=16$.}
\label{tab:model_specs}
\small
\setlength{\tabcolsep}{4pt}
\begin{tabular*}{\linewidth}{@{\extracolsep{\fill}}llcccc@{}}
\toprule
\textbf{Model} & \textbf{Family} & \textbf{Params} & \textbf{Type} & \textbf{Optimizer} & \stackhead{\textbf{Learning}\\\textbf{Rate}} \\
\midrule
BERT-base & BERT \citep{devlin2019bert} & 110M & Encoder (Full) & AdamW & $2{\times}10^{-5}$ \\
BERT-large & BERT \citep{devlin2019bert} & 340M & Encoder (Full) & AdamW & $2{\times}10^{-5}$ \\
RoBERTa-base & RoBERTa \citep{liu2019roberta} & 125M & Encoder (Full) & AdamW & $2{\times}10^{-5}$ \\
RoBERTa-large & RoBERTa \citep{liu2019roberta} & 355M & Encoder (Full) & AdamW & $2{\times}10^{-5}$ \\
DeBERTa-v3-base & DeBERTa \citep{he2021deberta} & 86M & Encoder (Full) & AdamW & $2{\times}10^{-5}$ \\
\midrule
Pythia-160M & Pythia \citep{biderman2023pythia} & 160M & Decoder (QLoRA) & AdamW & $2{\times}10^{-4}$ \\
Pythia-1.4B & Pythia \citep{biderman2023pythia} & 1.4B & Decoder (QLoRA) & AdamW & $2{\times}10^{-4}$ \\
Pythia-2.8B & Pythia \citep{biderman2023pythia} & 2.8B & Decoder (QLoRA) & AdamW & $2{\times}10^{-4}$ \\
Mistral-7B-v0.3 & Mistral \citep{jiang2023mistral} & 7.2B & Decoder (QLoRA) & AdamW & $2{\times}10^{-4}$ \\
Mistral-Nemo-12B & Mistral & 12.2B & Decoder (QLoRA) & AdamW & $2{\times}10^{-4}$ \\
Qwen3-14B & Qwen \citep{qwen2025qwen3} & 14.8B & Decoder (QLoRA) & AdamW & $2{\times}10^{-4}$ \\
\bottomrule
\end{tabular*}
\end{table}

\paragraph{Prompt and evaluation templates.}
For classification and hallucination detection on HaluEval \citep{li2023halueval}, input queries are structured with consistent instructions. Encoders concatenate input fields with standard separation tokens ($\texttt{[CLS]} \text{ Context } \texttt{[SEP]} \text{ Question } \texttt{[SEP]} \text{ Answer } \texttt{[SEP]}$). Decoders are evaluated using instruction-formatted prompts where the model outputs binary judgment tokens (\texttt{Yes} / \texttt{No} or \texttt{Hallucinated} / \texttt{Factual}) under greedy decoding.

\end{document}